# Multi-Cancer Detection Using a Computationally Efficient CNN with Transfer Learning

Vasileios E. Papageorgiou[1,3], Georgios Petmezas[1], Dimitrios-Panagiotis Papageorgiou[2], Leandros Stefanopoulos[4], Nicos Maglaveras[1]

**Affiliation**

[1]School of Medicine, Aristotle University of Thessaloniki, Thessaloniki, Greece
[2]Department of Mathematics, Aristotle University of Thessaloniki, Thessaloniki, Greece
[3]Department of Mathematics, National and Kapodistrian University of Athens, Athens, Greece
[4]Department of Electrical and Computer Engineering, Northwestern University, Evanston, IL, USA
Correspondence: vpapageor@math.auth.gr

## Abstract

This study introduces a computationally efficient convolutional neural network (CNN) architecture enhanced with transfer learning for multi-cancer detection using biomedical images. The proposed lightweight CNN model is designed to reduce computational complexity while maintaining high classification performance, making it suitable for deployment in resource-constrained environments. We evaluate this approach on three distinct tumor datasets comprising brain magnetic resonance imaging (MRI) and lung and kidney computed tomography (CT) scans. The model achieves test accuracy of 90.85 ± 2.22%, 98.64 ± 2.43% and 99.92 ± 0.08% for brain, lung, and kidney cancer classification, respectively, using 5-fold stratified cross-validation (CV). Transfer learning is employed by pretraining the model on one cancer type and fine-tuning it on the others, requiring only 20 additional epochs to achieve performance comparable to models trained from scratch. The fine-tuning process involves updating the classification part of the CNN and requires approximately 0.014 seconds per image per epoch using an NVIDIA GeForce GTX 960. Comparative evaluations show that the proposed model outperforms several state-of-the-art pretrained architectures, such as Xception, VGG16, VGG19, MobileNetV2 and DenseNet121. Overall, the model's effectiveness is evaluated across three types of cancer with distinct morphological characteristics, assessing its performance on both MRI and CT imaging modalities and demonstrating robust performance across diverse tasks and data types. These findings underscore the potential of streamlined deep learning (DL) frameworks in accelerating cancer diagnosis without sacrificing accuracy, especially in settings with limited computational resources.

## 1. Introduction

The use of artificial intelligence (AI) in medicine is expanding rapidly and increasingly impacting various facets of cancer care [1]. AI applications can be broadly classified into two categories: supervised and unsupervised learning [2-3]. In supervised learning, AI models are trained on labeled data to replicate human expertise in performing complex tasks with high accuracy and efficiency. In many real-world applications, such as image classification, speech recognition, or anomaly detection, early and accurate identification of critical patterns is essential for minimizing risk and improving outcomes [4].

A brain tumor or intracranial tumor is an abnormal mass of tissue in which cells grow unnaturally and uncontrollably. According to the World Health Organization (WHO), these types of tumors account for less than 2% of human cancers, although their severe morbidity and associated complications make early diagnosis a very important concept in modern medicine [5]. Intracranial tumors can be fatal, worsen the patient's standard of living and can affect men, women, or children.

Lung cancer ranks second, accounting for approximately 11.4% of all cancer cases, with an estimated 2.2 million cases worldwide. Lung cancer is the leading cause of death among other cancers, with deaths accounting for 18% of all cancers [6]. The survival rate of lung cancer patients 5 years after diagnosis ranges from 10% to 20%. Screening with CT can help detect lung cancer at an early stage so that the disease is more responsive to treatment [7]. In general, it has been reported that the patient's chances of living a long-life increase if the cancer is detected early, diagnosed and treated effectively [8].

Renal cell carcinoma (RCC) is the sixth most common cancer among all tumors in men and the tenth most common in women. Despite advances in understanding the molecular biology of RCC and refinements in therapies, treating patients with RCC at any stage of the disease is challenging. Diagnosis of early-stage renal cancer has improved significantly in recent decades with the widespread use of cross-sectional imaging [9]. Most renal carcinomas are initially detected as incidental renal masses on cross-sectional imaging (e.g., ultrasound, CT) performed for unspecified disease. Although most are discovered as small renal masses, earlier definitive therapeutic intervention for these tumors has not resulted in a significant improvement in cancer-specific mortality [10].

In the domain of medical imaging, several AI algorithms have been employed for brain cancer classification and detection. Notable examples include artificial neural networks (ANN) [11], K-nearest neighbors (KNN) [12] and support vector machines (SVM) [13-14]. However, convolutional neural networks (CNNs) are often considered the most effective approach for processing MRI and CT images due to their ability to capture complex spatial patterns in the data. A significant portion of the literature focuses on binary classification tasks, distinguishing between benign and malignant tumors. For instance, Babu et al. [15], Pathak et al. [16] and Kulkarni et al. [17] combine preprocessing and image magnification with traditional CNN methods to classify benign and malignant tumors. Saxena et al achieved 95% classification accuracy based on the ResNet50 architecture [18], albeit with obvious overfitting issues. As mentioned earlier, there are analyses that address multi-class classification tasks using a variety of preprocessing techniques and CNNs. The authors in [19-21] propose CNN architectures were used to classify images representing three types of brain tumors, namely pituitary adenomas, meningiomas and gliomas, from a public dataset, achieving accuracy of 94.39%, 90.89% and 98%, respectively. The study also compared classification performance across cropped, uncropped and segmented images. On the other hand, in [22], a Dense-Net-LSTM hybrid is used for a 4-class tumor classification problem, achieving 92.13% on a public database.

In studies on lung tumors, [23-24] the authors employed SVM classifiers and a CNN-based GoogLeNet model using the IQ-OTHNCCD dataset. In [23], the author preprocessed CT images using Gaussian filters, bit-plane slicing and image segmentation, achieving an accuracy of 89.88%. Meanwhile, in [24], they apply a Gabor filter and region of interest (ROI) extraction, obtaining an accuracy of 94.38%. Other machine learning (ML) approaches for lung cancer classification include the use of KNN [25], SVM [26], Naive Bayes [25] and Random Forests [27]. Polat and Mehr [28] propose a hybrid 3D CNN-SVM model, achieving a classification efficiency of 91.81%. Additionally, traditional CNN methods have been applied to private CT image datasets [29-30], incorporating techniques such as median filtering, contrast stretching and Otsu segmentation. Ardila et al. showed that an end-to-end 3D CNN can predict lung cancer risk from CT with state-of-the-art accuracy (AUC≈94.4%), underscoring the potential of CNNs in this domain [31].

Several studies have focused on renal cancer detection, particularly using preprocessed CT images. In ML approaches designed to distinguish between benign and malignant tumors [32-34], most research concentrated on differentiating low-fat angiomyolipomas from renal cancer, achieving promising results with AUC ranging from 90% to 96%. In contrast, Han et al. [35], employing a modified GoogLeNet architecture, observed lower performance (accuracy of 73%) in a three-class problem for identifying papillary RCCs (pRCCs), compared to the clearer distinction between clear cell RCCs (ccRCCs) and chromophobe RCCs (chrRCCs), where a binary classification approach yielded an accuracy of 85%. Xi et al. [36] reported a 70% accuracy using a ResNet-based ensemble model that combined preoperative T2-weighted and post-contrast T1 MRI sequences along with clinical variables to classify benign and malignant renal masses in 1162 patients. Additionally, Li et al. [37] successfully identified low- and high-grade ccRCC using MRI combined with patient history and radiologist-assigned imaging features, achieving an AUC of 0.845. Said et al. [38] implemented a random forest model based on features extracted from MRI images to perform binary classification of renal tumors.

Recent studies have shown the potential of transfer learning combined with deep convolutional neural networks (CNNs) in enhancing medical image classification tasks. Hassan et al. [39] utilized EfficientNetV2B0, a relatively deep architecture, for brain tumor classification, achieving 99.16% accuracy but relying on a computationally intensive model. Similarly, Kaur and Mahajan [40] applied ResNet152 to extract features from MRI. While effective, ResNet152 entails high computational cost and training complexity. Elmekki et al. [41] developed a two-stage transfer learning pipeline for cardiac ultrasound classification and grading, leveraging CNNs fine-tuned with large-scale architectures. Mugada and Lakshmi [42] proposed ATL-MobileNet for diabetic macular edema grading, incorporating an optimized MobileNet variant and a heuristic optimization algorithm, reflecting a growing effort toward lighter architectures, albeit still involving additional tuning overhead. Kim et al. [43]

demonstrated that transfer learning from domain-specific datasets like CAMELYON16 improves classification of frozen tissue sections. Their approach also depended on models pre-trained on large, diverse datasets and evaluated over complex whole-slide images. Finally, Kaur and Hans [44] in a comprehensive review noted that the majority of transfer learning applications in cancer diagnosis employ architectures like VGG, ResNet and Inception.

In contrast to these approaches, in this paper we highlight the capacity of computationally efficient CNN models to provide estimates of high precision to a multi-cancer task through transfer learning. The reduced complexity of this network makes it particularly suitable for working with small datasets, which are common in medical research, unlike many studies that rely on more complex and computationally intensive architectures. This approach not only reduces the risk of overfitting but also enhances flexibility, allowing the model to be retrained with minimal computational cost when new MRI or CT images are added to the dataset.

We evaluated the performance of the proposed methodology on three cancer datasets, which include brain MRI images, lung and kidney CT scans. In each experiment, we pretrained the CNN architecture on one dataset and then fine-tuned it on the other two, resulting in notable improvements in classification performance. It is important to note that fine-tuning affected only the classification part of the CNN. Moreover, particular attention was given to identifying the optimal number of epochs required to achieve maximum detection performance. The results demonstrate that low-complexity models can deliver robust predictions for real-world applications and are well-suited for transfer learning tasks with minimal computational resources, highlighting their generalizability.

Transfer learning is particularly necessary in medical imaging because labeled datasets are often limited, expensive to curate, and difficult to balance across classes. By initializing a model with knowledge learned from a related source task, transfer learning enables faster convergence, reduces computational cost, and often improves generalization compared with training from scratch. This is especially important in cancer detection problems, where robust feature extraction must be achieved despite restricted data availability. Therefore, transfer learning provides a practical and effective strategy for building accurate diagnostic models while minimizing training requirements.

Although the datasets used in this study are moderate in size, computational efficiency remains important for several reasons. First, medical imaging models are rarely trained only once. In practice, they are repeatedly retrained during cross-validation, hyperparameter tuning, and model adaptation to new institutions, scanners, or cancer types. Second, many research and clinical environments do not have access to high-end GPU infrastructure, making lightweight models more practical for routine use. Third, in transfer learning settings such as ours, efficiency is especially relevant because the same architecture is pretrained on one dataset and then fine-tuned multiple times on other target datasets. Therefore, even when each individual dataset is not extremely large, the cumulative computational cost can become substantial. For this reason, reducing model complexity is important not only for faster inference, but also for feasible retraining, reproducibility, and broader real-world deployment.

## 2. Materials and Methods

In this section, we present a computationally efficient CNN model together with the transfer learning process. We focus on key concepts related to CNN architectures, the operation of the Adam optimizer, and the principles of transfer learning. Additionally, we provide the formulas for the four evaluation measures used and outline the relevant details of the brain MRI, lung, and kidney CT scan datasets.

### 2.1 CNN models

CNNs are a type of supervised learning typically based on convolutional kernels and are primarily used for visual tasks such as video analysis, image segmentation and classification tasks. A CNN consists of several key components: convolutional layers, pooling layers, batch normalization layers and fully connected layers. These layers are arranged in a specific order, with feature extraction layers followed by fully connected layers to produce the final output [45].

Convolutional layers consist of multiple kernels which are the layers' trainable parameters, modified during each iteration. Let $\boldsymbol{X}^k \in \mathbb{R}^{M^k \times N^k \times D^k}$ be the input of the $k$–th convolutional layer and $\boldsymbol{F} \in \mathbb{R}^{m \times n \times D^k \times S}$ be a four-dimensional tensor representing the $S$ kernels of $k$–th layer. Each kernel is of spatial span $m \times n$. Considering the simple case where stride is 1 and no padding is used, the output of the $k$–th convolutional layer is a three-dimensional tensor denoted as $\boldsymbol{Y}^k \in \mathbb{R}^{(M^k - m + 1) \times (N^k - n + 1) \times S}$. The elements of this tensor result from the equation

$$y_{i^k, j^k, s} = \sum_{i=0}^{m-1} \sum_{j=0}^{n-1} \sum_{l=0}^{D^k - 1} F_{i,j,l,s} \times x^k_{i^k + i, j^k + j, l}. \tag{1}$$

Equation (1) is repeated for all $0 \leq s < S$, and for any spatial location satisfying $0 \leq i^k < M^k - m + 1$ and $0 \leq j^k < N^k - n + 1$ [46].

Let $\boldsymbol{X}^k \in \mathbb{R}^{M^k \times N^k \times D^k}$ be the input of the $k$–th layer that is now a pooling layer with a spatial span of $m \times n$. We assume that $m$ divides $M$ and $n$ divides $N$ and the stride equals the selected spatial span. The output is a three-dimensional tensor $\boldsymbol{Y}^k \in \mathbb{R}^{M^{k+1} \times N^{k+1} \times D^{k+1}}$, where

$$M^{k+1} = \frac{M^k}{m}, \quad N^{k+1} = \frac{N^k}{n}, \quad D^{k+1} = D^k, \tag{2}$$

while the pooling layer operates upon $\boldsymbol{X}^k$ channel by channel independently. In our network we utilize two max pooling layers. The elements of the output of a max pooling are given by

$$y_{i^k, j^k, d} = \max_{0 \leq i < m, 0 \leq j < n} x^k_{i^k \times m + i, j^k \times n + j, d}, \tag{3}$$

where $0 \leq i^k < M^k$, $0 \leq j^k < N^k$and $0 \leq d < D^k$. These layers reduce the tensor's dimensions while preserving the important detected patterns [47].

The fully connected layers form the second part of a CNN and are used to capture the most relevant features extracted by the convolutional layers. The input of the first fully connected layer is a high-dimensional vector containing all extracted features produced by a flattening operation. Finally, techniques widely used to introduce nonlinearity and improve training stability include the ReLU activation function and batch normalization. The ReLU function is given by

$$y_{i,j,d} = max\left(0, x_{i,j,d}^{k}\right), \tag{4}$$

with $0 \leq i < M^k$, $0 \leq j < N^k$ and $0 \leq d < D^k$, aiming to transfer only the informative elements for the classification.

A grayscale medical image $\boldsymbol{X}$ can be represented as a matrix, formally defined as $\boldsymbol{X} \in \mathbb{R}^{M \times N}$, where $M$ and $N$ are the height and the width of the image respectively. The input matrix corresponding to $i$ −th medical image passes through successive layers to produce the prediction $\hat{y}_i$. An error value is then computed using a predefined loss function. In most cases, the Cross-Entropy loss denoted as $L_{CE}$, is used. For binary classification scenarios, the Binary Cross-Entropy loss function ($L_{BCE}$) is applied. Given a sample of $n$ medical images, it is given by

$$L_{BCE}(\boldsymbol{y}, \hat{\boldsymbol{y}}) = -\frac{1}{n}\sum_{i=1}^{n}\left(y_i log(\hat{y}_i) + (1 - y_i) log(1 - \hat{y}_i)\right) , \tag{5}$$

where $y_i \in \{0, 1\}$ corresponds to the image's ground truth.

### 2.2. Optimizer

After each forward pass of the neural network, the computed error is used in the learning process to update the network's trainable parameters based on an optimization algorithm. Let $\boldsymbol{\theta}_t$ denote the vector of parameters at time step $t$ and let $\eta$ represent the learning rate of the optimization procedure. According to gradient descent

$$\boldsymbol{\theta}_{t+1} = \boldsymbol{\theta}_t - \eta \, \nabla_{\boldsymbol{\theta}} L_{BCE}(\boldsymbol{\theta}_t) , \tag{6}$$

where $\nabla_{\boldsymbol{\theta}} L_{BCE}(\boldsymbol{\theta}_t)$ denotes the gradient of the binary cross-entropy loss function evaluated at $\boldsymbol{\theta}_t$.

In Adam, the parameter update process is modified by introducing two key components, namely the first (momentum) and the second moment estimate (uncentered variance) [48]. An exponentially decaying average of the past gradients is used. The first moment estimate is computed iteratively based on

$$\boldsymbol{m}_t = \alpha_1 \boldsymbol{m}_{t-1} + (1 - \alpha_1) \nabla_{\boldsymbol{\theta}} L_{BCE}(\boldsymbol{\theta}_t) , \tag{7}$$

where $\alpha_1$ is a hyperparameter controlling the decay rate. The default value is $\alpha_1 = 0.9$, while values close to 1 are usually preferable.

We also maintain an exponentially decaying average of the past squared gradients (uncentered variance). This is the second moment estimate, denoted by $\boldsymbol{v}_\tau$

$$\boldsymbol{v}_t = a_2 \boldsymbol{v}_{t-1} + (1 - a_2)\left(\nabla_{\boldsymbol{\theta}} L_{BCE}(\boldsymbol{\theta}_t)\right)^2, \tag{8}$$

where $a_2$ takes values close to unity with the default value being $a_2 = 0.999$.

Since $\boldsymbol{m}_t$ and $\boldsymbol{v}_t$ are initialized as zero, their estimates are biased towards this value during the initial time steps. To address this issue, the following bias-corrected versions are used, i.e.

$$\widetilde{\boldsymbol{m}}_t = \frac{\boldsymbol{m}_t}{1 - \alpha_1^t}, \tag{9}$$

and

$$\widetilde{\boldsymbol{v}}_t = \frac{\boldsymbol{v}_t}{1 - \alpha_2^t}. \tag{10}$$

This correction ensures that $\widetilde{\boldsymbol{m}}_t$ and $\widetilde{\boldsymbol{v}}_t$ are unbiased estimates of the first and second moments, respectively. Finally, the parameter update rule in Adam is

$$\boldsymbol{\theta}_{t+1} = \boldsymbol{\theta}_t - \eta \frac{\widetilde{\boldsymbol{m}}_t}{\sqrt{\widetilde{\boldsymbol{v}}_t} + \varepsilon}, \tag{11}$$

where $\varepsilon \in \mathbb{R}$ is a small constant (e.g., $10^{-8}$) added to prevent division by zero.

## 2.3 Transfer Learning

Transfer learning is an ML technique where a model developed for a specific task is reused as the starting point for a model on a second task. The core idea behind transfer learning is to leverage the knowledge gained from solving one problem and apply it to a different, often related problem. Mathematically, transfer learning can be framed as a process of optimizing a function where the initial model parameters (learned from the source task) are used as a starting point for the optimization process in the target task.

Let $L_T(\boldsymbol{\theta})$ be the loss function of the target task, where $\boldsymbol{\theta}$ denotes the model's parameters. In transfer learning, we typically initialize the model using the weights $\boldsymbol{\theta}_0$ from the pretrained model. This can be viewed as the solution to an optimization problem for the source task

$$\boldsymbol{\theta}_0 = arg \min_{\boldsymbol{\theta}} L_S(\boldsymbol{\theta}), \tag{12}$$

where $L_S(\boldsymbol{\theta})$ is the loss function of the source task. When transferring this model to the target task, we fine-tune its weights $\boldsymbol{\theta}$ by minimizing the target task's loss function

$$\boldsymbol{\theta}^* = arg \min_{\boldsymbol{\theta}} L_T(\boldsymbol{\theta}), \tag{13}$$

using gradient-based optimization, while $\boldsymbol{\theta}^*$ denotes the final parameters after fine-tuning. The weights for the target task are initialized using $\boldsymbol{\theta}_0$.

The idea is that the learned representations (features) from the source task are generally useful for the target task, especially if both tasks are related. Hence, the optimization for the target task can be performed more efficiently compared to training a model from scratch. In our case we train the low-complexity CNN model on the source task for 75 epochs using a learning rate of $\eta = 10^{-3}$. After receiving the optimal set of parameters ($\boldsymbol{\theta}_0$) we retrain the acquired model for a smaller number of epochs using one-tenth of the initial learning rate. In the Results section we demonstrate the efficacy of our approach and present an extensive analysis aimed at identifying the optimal number of epochs during the retraining phase.

### 2.4 Evaluation Measures

In the model evaluation stage, we assess the performance of the trained model using well-established metrics commonly used for classification tasks. These are accuracy, sensitivity, specificity and the F1 score. These are formally defined as

$$Accuracy = \frac{TP + TN}{TP + TN + FP + FN}, \tag{14}$$

$$Sensitivity = \frac{TP}{TP + FN}, \tag{15}$$

$$Specificity = \frac{TN}{TN + FP}, \tag{16}$$

and

$$F1\ score = \frac{2TP}{2TP + FP + FN}. \tag{17}$$

Accuracy denotes the percentage of the correctly classified cases from the trained model and is the simplest and intuitive metric. As a general measure, F1 score is the harmonic average between precision and sensitivity, and high scores reflect an equally balanced model of both tumorous and non-tumorous instances.

### 2.5 Datasets

The brain cancer dataset consists of 3000 images, which can be used for both training and testing the proposed CNN model. The dataset is well-balanced, with 1500 images for normal cases and 1500 for tumorous cases. It is publicly available and has been uploaded to Kaggle [49].

The kidney cancer dataset was sourced from a broader collection of cancer-related datasets, which includes cases from various types of cancer, such as brain and breast cancer. It consists of 10000 images, with 5,000 normal and 5,000 tumorous images, resulting in a completely balanced dataset [50].

The lung cancer dataset was collected by specialists at the Iraq Oncology Teaching Hospital/National Center for Cancer Diseases (IQ-OTH/NCCD) over a three-month period in 2019. It comprises 1190 CT scan images from 110 patients: 40 with malignant tumors, 15 with benign tumors and 55 healthy (normal) cases. Since the task is a binary classification problem focused on detecting tumorous regions, both benign and malignant cases are labeled as tumorous [51]. Figure 1 shows original and cropped instances of brain, kidney and lung images.

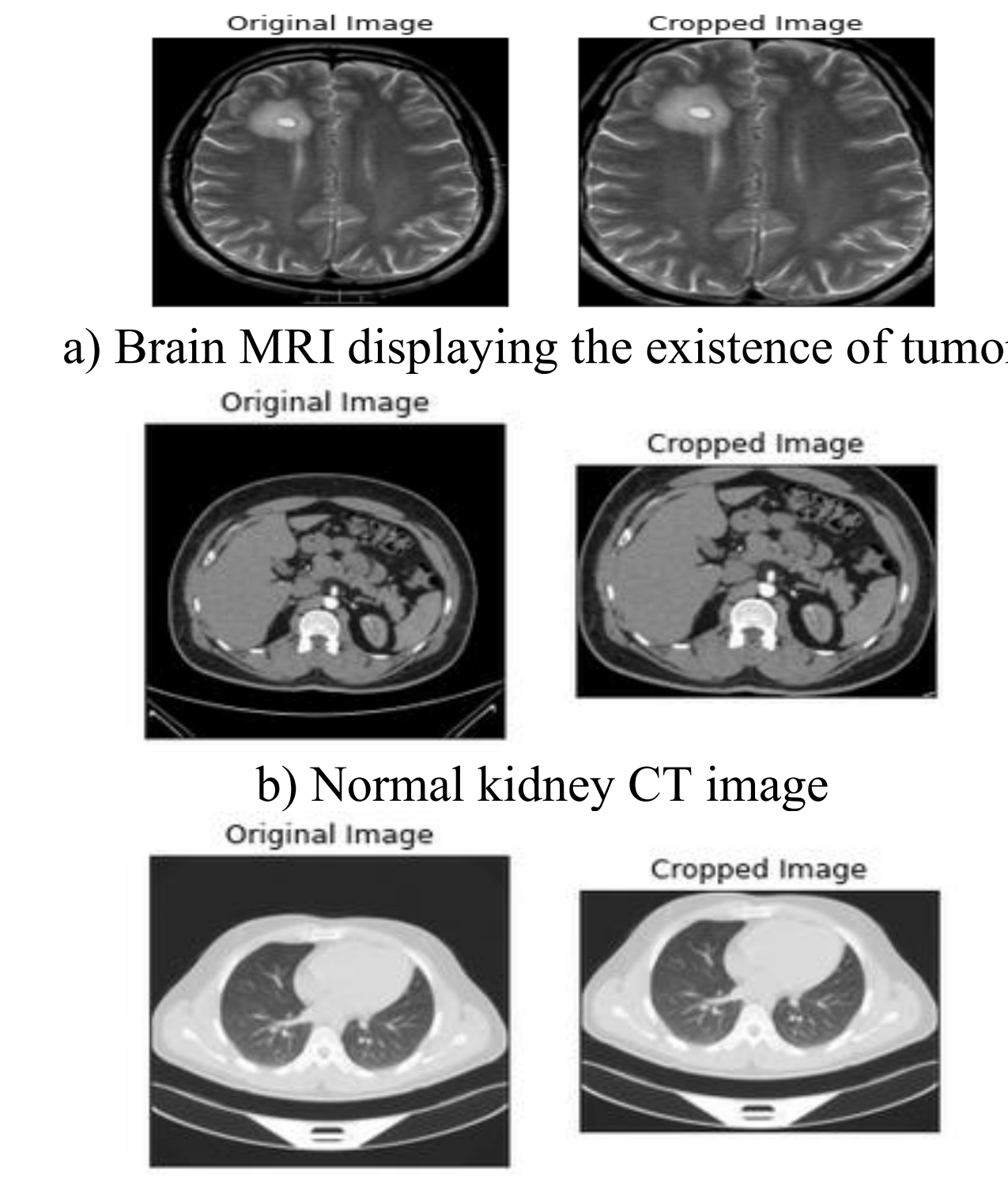

a) Brain MRI displaying the existence of tumor

b) Normal kidney CT image

c) Normal lung CT image

**Figure 1.** Original and cropped MRI and CT images

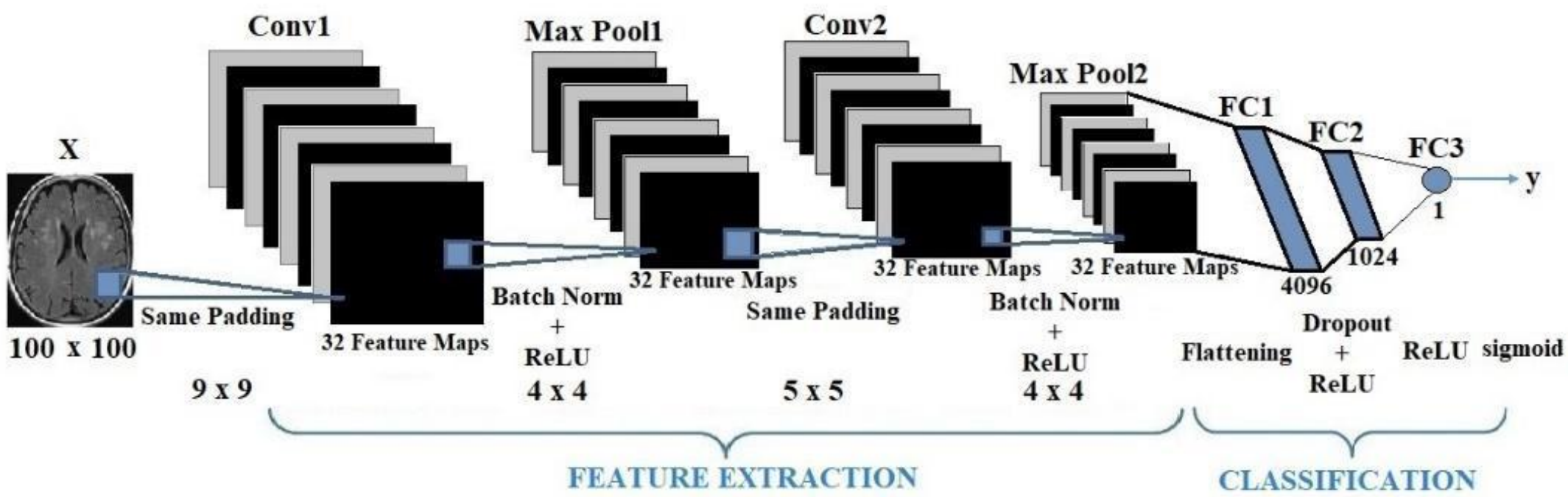

**Figure 2.** Representation of the computationally efficient CNN architecture

## 3. Results

### 3.1 CNN Structure and Pretraining Process

In this section, we introduce a low-complexity CNN architecture consisting of seven main layers. This architecture was selected based on experimental evaluation [52]. Reducing the number of layers led to decreased performance, while adding additional convolutional or fully connected layers did not yield any improvement. Therefore, the proposed structure offers the best balance between accuracy and computational efficiency. This architecture has shown satisfactory performance even in tasks like the classification of cardiac arrhythmia [53].

The first four layers, two convolutional followed by two max pooling layers, are responsible for feature extraction. The remaining three fully connected layers leverage the extracted features to achieve high classification performance (see Figure 2). The model accepts 2D grayscale input images of size 100×100. This small input size helps reduce the computational load while maintaining the necessary information of the input images. The architecture begins with a convolutional layer using 32 kernels of size 9×9, followed by a max pooling layer with a 4×4 window. This structure is repeated, with a 5×5 convolutional layer and a 4×4 max pooling layer. In both cases padding is applied before the convolution, while ReLU activation functions and batch normalization are used for each layer. The second part of the architecture consists of three fully connected and two dropout layers (at a rate of 0.2) to mitigate overfitting.

The proposed low-complexity scheme is employed to investigate the overall classification efficiency on all three datasets containing lung, kidney and brain cancer images. For each dataset we utilized a stratified 5-fold CV splitting strategy. Images of the same patient were not included in both the training and test sets. Each time 3 parts are used for training the CNN architecture and the other 2 parts for the validation and test process. Thus, we result in test sets containing 600, 2000 and 238 images for the brain, kidney and lung datasets correspondingly. Several learning rates were utilized during the training phase, namely $\eta \in \{10^{-1}, 5 \times 10^{-2}, 10^{-2}, 5 \times 10^{-3}, 10^{-3}, 5 \times 10^{-4}, 10^{-4}\}$, while the best classification results were generated for $\eta = 10^{-3}$. We trained our model for 75 epochs using the Adam optimizer on an NVIDIA GeForce GTX 960 GPU. For reproducibility purposes, Table 1 summarizes the main implementation details of the proposed pipeline, including preprocessing, input configuration, cross-validation strategy, and the fine-tuning setup used for both the proposed CNN and the comparison pretrained models. Details of the table referring to the fine-tuning process are also included in Section 2.

**Table 1.** Summary of the implementation pipeline and reproducibility details

| Component | Proposed low-complexity CNN |
|---|---|
| Input image type | 2D grayscale images |
| Cropping | ROI was identified based on grayscale intensity contrast and background removal, and the image was cropped to the corresponding bounding region. |
| Resizing | After cropping each image was resized to 100 × 100 pixels. |
| Preprocessing consistency | The proposed CNN was trained on grayscale images |
| Data splitting strategy | Stratified 5-fold CV |
| Patient-level separation | Images from the same patient were not included in both training and test sets |
| Fold usage | Three parts were used for training and 2 parts for validation and test |
| Test set size | Brain: 600 images; Kidney: 2000 images; Lung: 238 images |
| CNN architecture | Two convolutional layers, two max-pooling layers, and three fully connected layers |
| Convolution blocks | First convolution: 32 kernels of size 9 × 9; second convolution: kernels of size 5 × 5; padding, batch normalization and ReLU were applied |
| Pooling | Two max-pooling layers with 4 × 4 window |
| Regularization | Two dropout layers with dropout rate 0.2 |
| Batch size | 32 |
| Optimizer | Adam |
| Initial training epochs | 75 epochs |
| Fine-tuning epochs | 5, 10, 15, 20 and 25 epochs were examined |
| Fine-tuning learning rate | $10^{-4}$ |
| Fine-tuned parameters | Only the classification part of the CNN |
| Hardware | NVIDIA GeForce GTX 960 GPU |

## 3.2 Pretrained CNN using Brain MRI images

After the training of the CNN described above on the brain MRI tumor database for 75 epochs with a batch size of 32, we result in Table 2 which shows the training, validation and test performance of the introduced scheme. In each cell we observe the average performance followed by the standard deviation of the classification accuracy, specificity, sensitivity and F1 score acquired from the 5-fold CV. The overall training procedure, considering all 5 folds and 75 epochs lasted for about 2002 seconds. According to Table 2, our model captures efficiently the patterns of brain MRI images, with an almost perfect training process, reaching an accuracy of 98.74 ± 0.71%, specificity of 98.58 ± 0.65%, sensitivity of 99.08 ± 0.45% and F1 score of 98.77 ± 0.60%.

Satisfactory performance is also observed regarding the test set of the brain tumor dataset. The introduced scheme achieves an accuracy of 90.85 ± 2.22%, specificity of 85.91 ± 7.09%, sensitivity of 95.20 ± 2.01% and F1 score of 91.70 ± 2.32%. We note that the results corresponding to the test subset are not as balanced as those accompanying the validation process. Specifically, the mean deviation between specificity and sensitivity is 9.29 units, while the deviation in the validation process is 3.8 units. As a result, the introduced model seems to "better understand" the patterns of tumorous MRI, while some non-tumorous images may have been misinterpreted as tumorous. Fortunately, the Type I error (false positive) in tumor detection is generally less critical than a Type II error (false negative), as a false alarm can often be resolved through additional medical examinations, whereas missing a tumor could have severe consequences.

**Table 2.** Classification performance of the low-complexity CNN based on the brain MRI dataset

| | **Accuracy (%)** | **Specificity (%)** | **Sensitivity (%)** | **F1 score (%)** |
|---|---|---|---|---|
| **Training Set** | 98.74 ± 0.71 | 98.58 ± 0.65 | 99.08 ± 0.45 | 98.77 ± 0.60 |
| **Validation Set** | 90.93 ± 1.67 | 89.13 ± 1.71 | 92.93 ± 2.80 | 90.99 ± 1.73 |
| **Test Set** | 90.85 ± 2.22 | 85.91 ± 7.09 | 95.20 ± 2.01 | 91.70 ± 2.32 |

After evaluating the classification efficacy of the proposed CNN architecture on the brain tumor dataset, we proceeded to experiment with the effectiveness of the transfer learning procedure. We take the already trained network and retrain it for 5, 10, 15, 20 and 25 epochs using one-tenth of the original learning rate. Figures 3 and 4 exhibit the evolution of the four selected evaluation measures for the abovementioned number of epochs considering the kidney and lung CT datasets, respectively. The error bars display the magnitude of the standard deviation produced by the 5-fold CV. Each 5-epoch fine-tuning cycle required approximately 134 seconds. It should be noted that the fine-tuning process involves retraining the classification part of the CNN, specifically updating the weights of the fully connected layers.

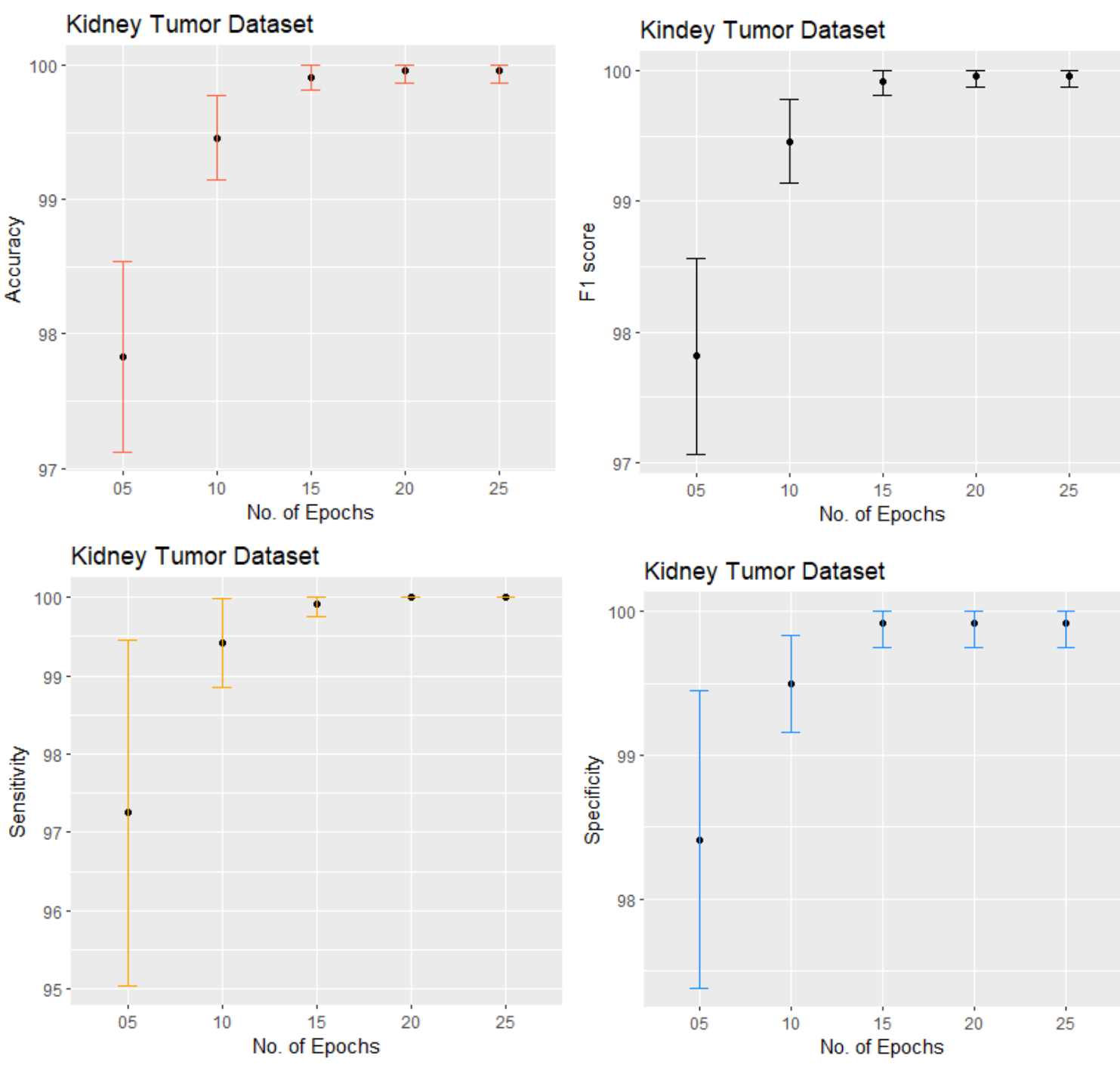

**Figure 3.** Kidney tumor detection based on the brain MRI pretrained CNN model

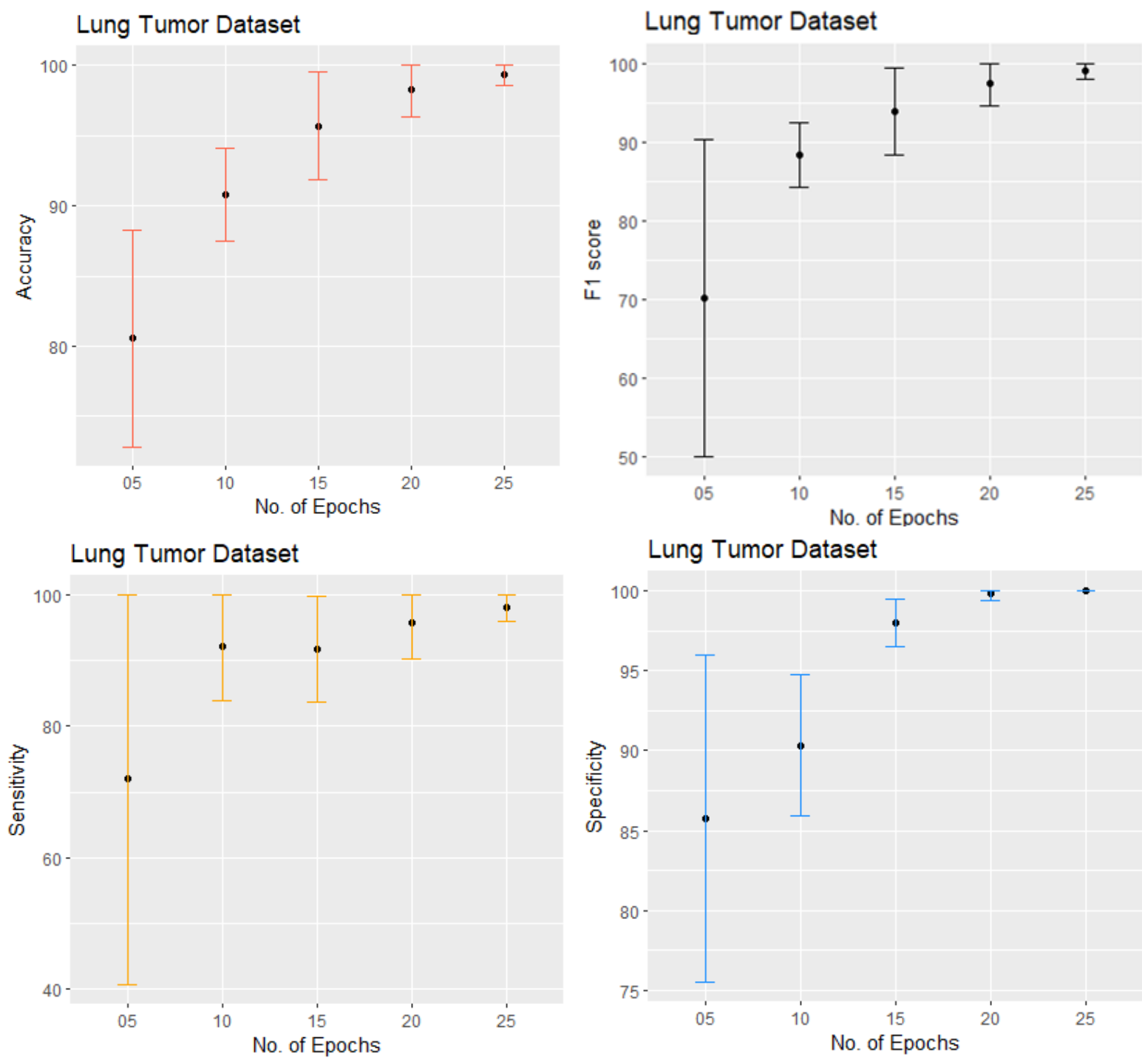

**Figure 4.** Lung tumor detection based on the brain MRI pretrained CNN model

According to Figure 3, the retrained model demonstrates increasing performance with the number of epochs, reaching its peak at 20 epochs. Additionally, the deviations from the mean estimate decrease noticeably as the number of epochs increases. The model's performance remains effectively unchanged between 20 and 25 epochs, suggesting that the additional 5 epochs may be redundant. In contrast, for the lung CT scan dataset, classificationperformance increases monotonically, accompanied by a consistent reduction in error intervals. However, the gain in accuracy from 20 to 25 epochs, from 98.6% to 99.1%, is relatively modest. For both the kidney and lung datasets, the deviations between sensitivity and specificity are minimal, indicating a robust retraining process. Compared with the train-from-scratch performance on the brain tumor test set (accuracy 90.85%), the brain-pretrained model achieved markedly higher performance after fine-tuning on the kidney and lung datasets, reaching accuracies of 99.96% and 98.45%, respectively, thereby demonstrating the effectiveness and transferability of the learned representations.

### 3.3 Lung CT scan pretrained CNN

For the second experiment, we trained the computationally efficient CNN architecture on the lung CT scan dataset for 75 epochs with a batch size of 32. Table 3 summarizes the performance of the proposed model across training, validation and test phases. Each cell in the table presents the average classification accuracy, specificity, sensitivity and F1 score, along with their corresponding standard deviations, which were obtained through a 5-fold CV. The total time taken for the entire training process considering all 5 folds and 75 epochs was approximately 945 seconds. The training time required is shorter compared to the running time observed with the brain tumor dataset, which is anticipated, as the lung cancer dataset has fewer CT instances.

The model effectively learns the patterns of lung CT scans, achieving near-perfect results in training and validation accuracies 97.98 ± 1.47% and 97.88 ± 1.26%, specificities 96.85 ± 1.73% and 92.00 ± 7.55%, sensitivities 98.25 ± 1.41% and 98.89 ± 1.16% and F1 scores 98.05 ± 1.36% and 97.75 ± 1.28%, correspondingly. The performance on the test set also demonstrates satisfactory results. The model achieved an accuracy of 98.64 ± 2.43%, specificity of 99.64 ± 0.33%, sensitivity of 96.99 ± 5.25% and F1 score of 98.12 ± 3.37%. Overall, the lung cancer detection task appears to be less demanding for our model, as we observe better performance compared to the brain MRI dataset.

**Table 3.** Classification performance of the low-complexity CNN based on the lung CT scans dataset

| | Accuracy (%) | Specificity (%) | Sensitivity (%) | F1 score (%) |
|---|---|---|---|---|
| **Training Set** | 97.98 ± 1.47 | 96.85 ± 1.73 | 98.25 ± 1.41 | 98.05 ± 1.36 |
| **Validation Set** | 97.88 ± 1.26 | 92.00 ± 7.55 | 98.89 ± 1.16 | 97.75 ± 1.28 |
| **Test Set** | 98.64 ± 2.43 | 99.64 ± 0.33 | 96.99 ± 5.25 | 98.12 ± 3.37 |

After evaluating the classification performance of the proposed CNN architecture on the lung cancer CT scans, we retrained our model on the brain and kidney tumor datasets independently, for 5, 10, 15, 20 and 25 epochs using one-tenth of the original learning rate. Figures 5 and 6 show the progression of the four key evaluation measures for the brain and kidney tumor datasets, respectively. Once again, we note that narrower error bars indicate a model that delivers more consistent estimations.

In Figure 5, we observe a unimodal increase in performance measures for the brain MRI dataset, with maximum performance reached at 20 epochs. Specifically, after retraining for 20 epochs, the model achieves an average accuracy of 89.88%, specificity of 84.84%, sensitivity of 94.92% and an F1 score of 90.38%. These performance metrics decline when the model is trained for an additional 5 epochs. Moreover, training for 20 epochs offers an advantage over 25 epochs, as it yields smaller deviations from the mean. Therefore, 20 epochs represent the most appropriate choice for this dataset in terms of both predictive efficiency and computational cost. The deterioration in classification performance is attributed to the occurrence of overfitting issues, validating the necessity of the numerical experimentation with respect to the number of epochs associated with the fine-tuning process.

In Figure 6, we see that the optimal predictive performance occurs after retraining for 15 epochs, following a unimodal trend. The model's performance declines at 20 and 25 epochs, rendering the additional 5 and 10 epochs unnecessary. Specifically, after retraining for 15 epochs the model achieves a mean accuracy of 99.87%, specificity of 99.75%, sensitivity of 100% and an F1 score of 99.87%. It should be noted that as the model's performance improves, the consistency of its estimations also increases, with the error bars narrowing and closely aligning with the average classification estimate. Compared with the train-from-scratch performance on the lung CT test set (accuracy 98.64%), the lung-pretrained model achieved lower performance after fine-tuning on the brain tumor dataset (accuracy 89.88%) but even higher performance on the kidney dataset (accuracy 99.87%), indicating that the transferability of the learned representations depends on the similarity and complexity of the target task.

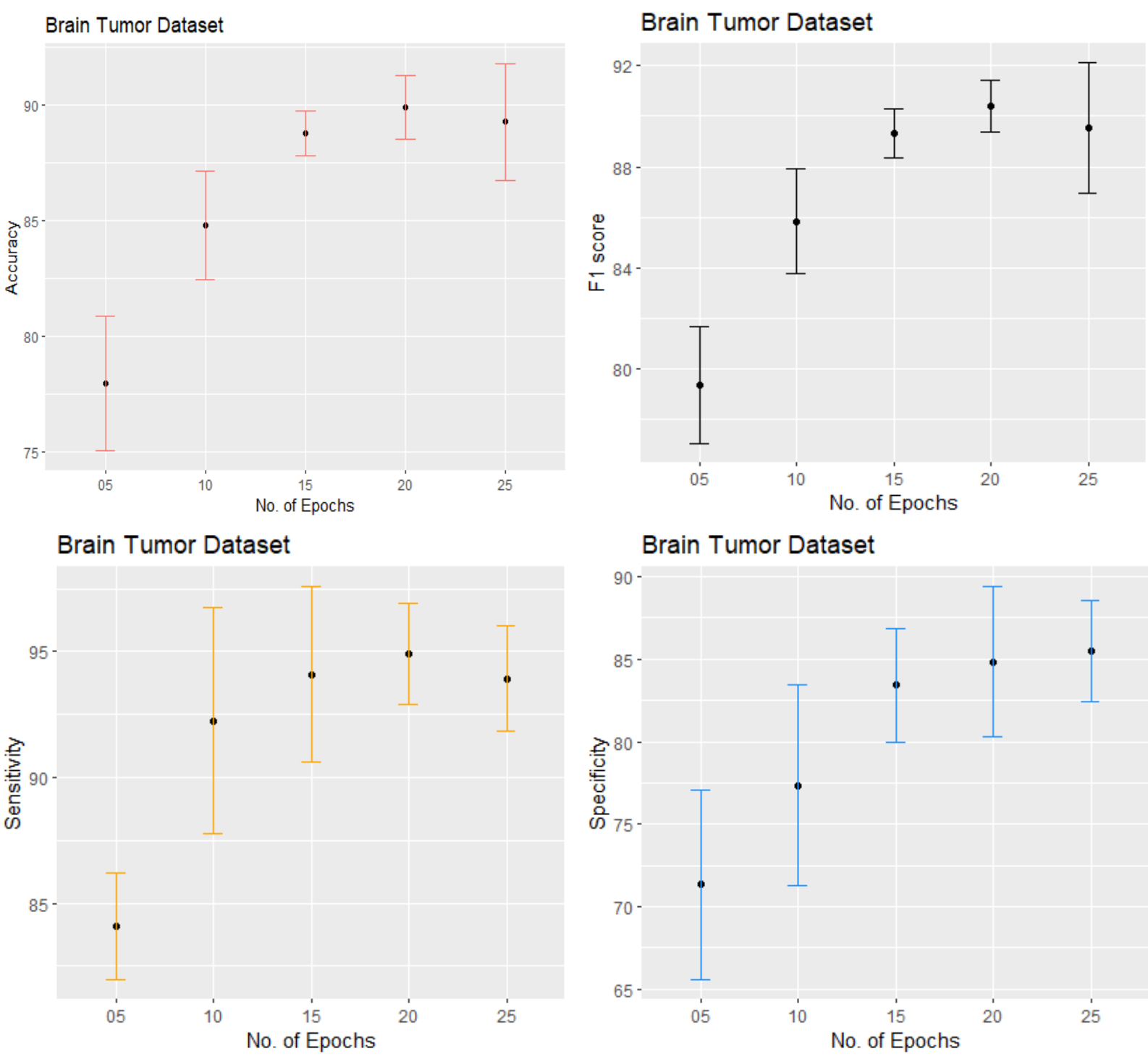

**Figure 5.** Brain tumor detection based on the lung CT pretrained CNN model

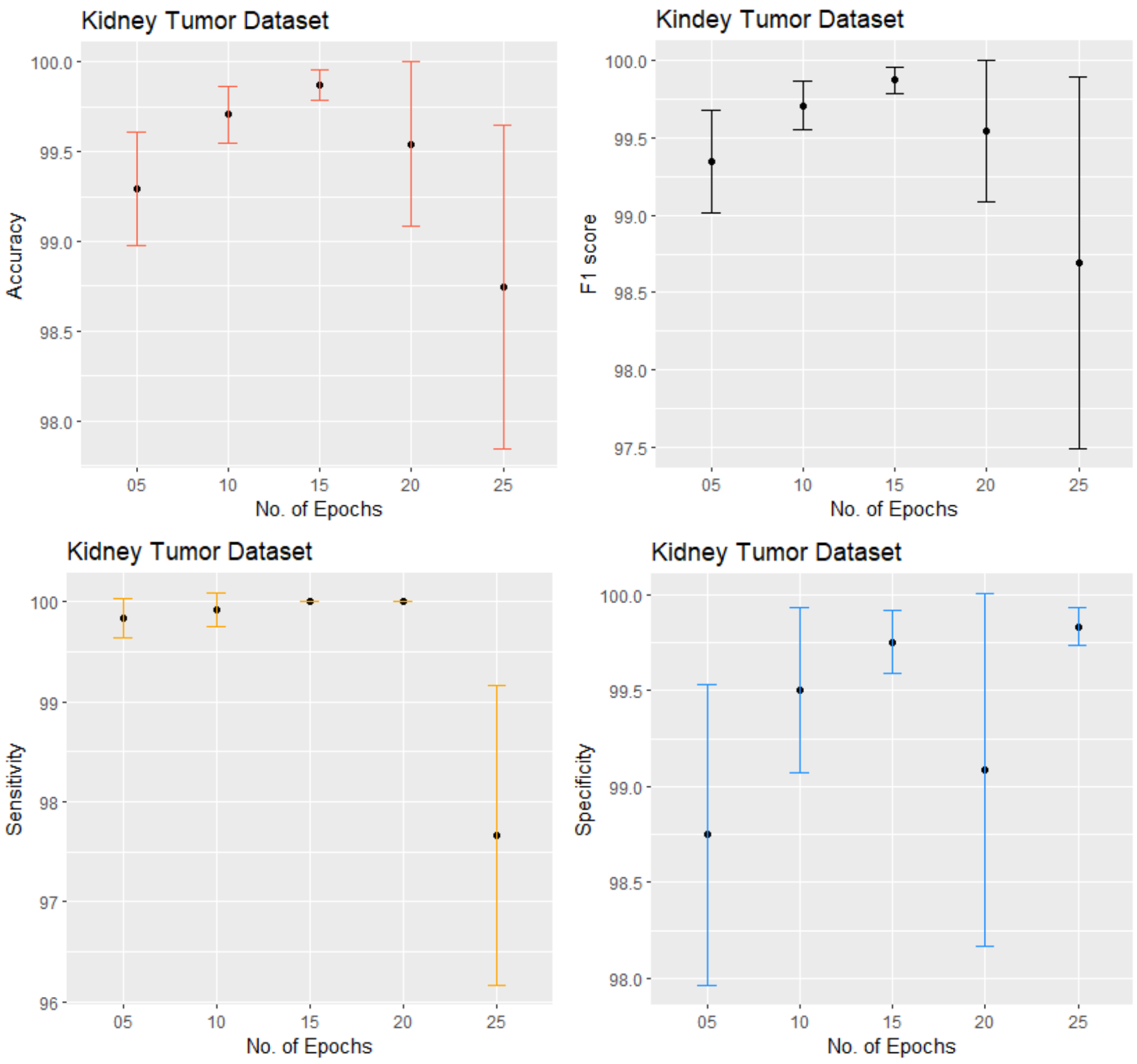

**Figure 6.** Kidney tumor detection based on the lung CT pretrained CNN model

### 3.4 Kidney CT scan pretrained CNN

The source task in the third experiment involves training the CNN on kidney CT scans, while the target tasks are the retraining on brain MRI and lung CT images. In Table 4 we show an overview of the model's performance during the training, validation and test process. The total training time including all 5 folds and 75 epochs, was approximately 6410 seconds, considering that the training set contains 6000 images for every cross-validation iteration. The model demonstrated an almost perfect training, validation and test classification performance. We obtained a training accuracy of 99.99 ± 0.01%, a specificity of 99.97 ± 0.03%, a sensitivity of 99.97 ± 0.03%, and an F1 score of 99.98 ± 0.02%. The model's performance on the test set yielded promising results, with an accuracy of 99.92 ± 0.08%, specificity of 99.83 ± 0.17%, sensitivity of 100 ± 0.00% and F1 score of 99.96 ± 0.08%. The identification of non-tumorous and tumorous instances is well-balanced as the deviation between specificity and sensitivity remains below unity across all three phases.

**Table 4.** Classification performance of the low-complexity CNN based on the kidney CT scans dataset

| | Accuracy (%) | Specificity (%) | Sensitivity (%) | F1 score (%) |
|---|---|---|---|---|
| **Training Set** | 99.99 ± 0.01 | 99.97 ± 0.03 | 99.97 ± 0.03 | 99.98 ± 0.02 |
| **Validation Set** | 99.58 ± 0.22 | 99.79 ± 0.25 | 99.37 ± 0.59 | 99.51 ± 0.21 |
| **Test Set** | 99.92 ± 0.08 | 99.83 ± 0.17 | 100.00 ± 0.00 | 99.96 ± 0.08 |

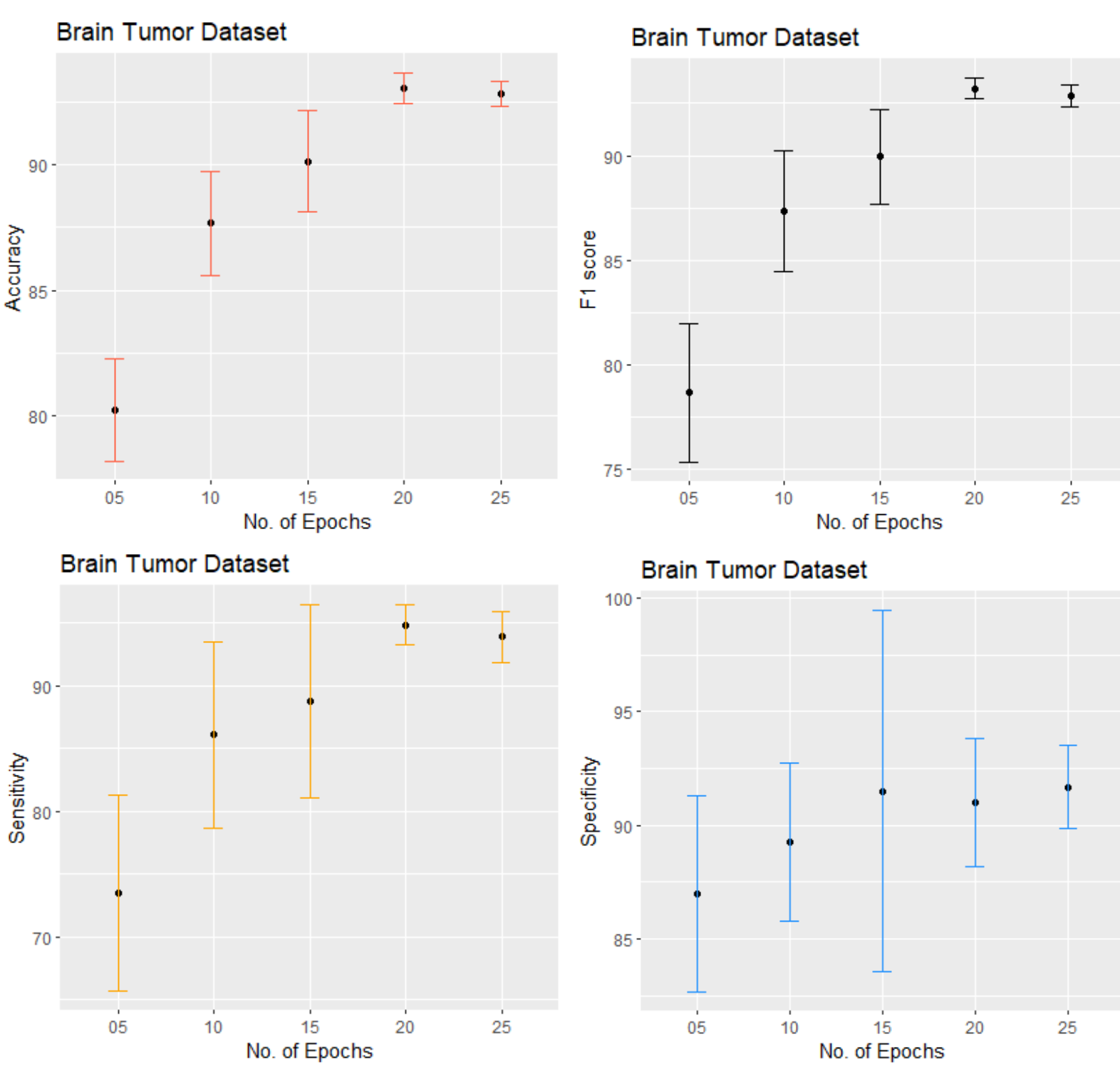

**Figure 7.** Brain tumor detection based on the kidney CT pretrained CNN model

After investigating the predictive efficacy of the CNN on the kidney CT scans, we proceed to retrain the model on the brain and lung cancer datasets, independently. For each 5-epoch retraining process, a runtime of about 47 and 21 seconds was required. In Figures 7 and 8, a unimodal performance trend is observed for both the brain MRI and lung CT scan datasets, with peak performance reached after 20 epochs. Performance declines when the model is trained for an additional 5 epochs, rendering the extra training unnecessary. Furthermore, using 20 epochs results in satisfactorily consistent performance estimates.

Furthermore, based on Figure 7 and using 20 epochs, the model retrained on the brain tumor dataset achieves an average accuracy of 92.54%, specificity of 91.02%, sensitivity of 94.87% and F1 score of 92.71%. Finally, as shown in Figure 8, which illustrates the predictive efficiency of the retrained model on lung CT scans, training for 20 epochs yields a mean accuracy of 99.18%, specificity of 99.56%, sensitivity of 98.55% and F1 score of 99.10%. Compared with the train-from-scratch performance on the kidney CT test set (accuracy 99.92%), the kidney-pretrained model maintained high performance after fine-tuning on the lung dataset (accuracy 99.18%) and achieved substantially lower but still competitive performance on the brain tumor dataset (accuracy 92.54%), further indicating that transferability depends on the characteristics and difficulty of the target imaging task.

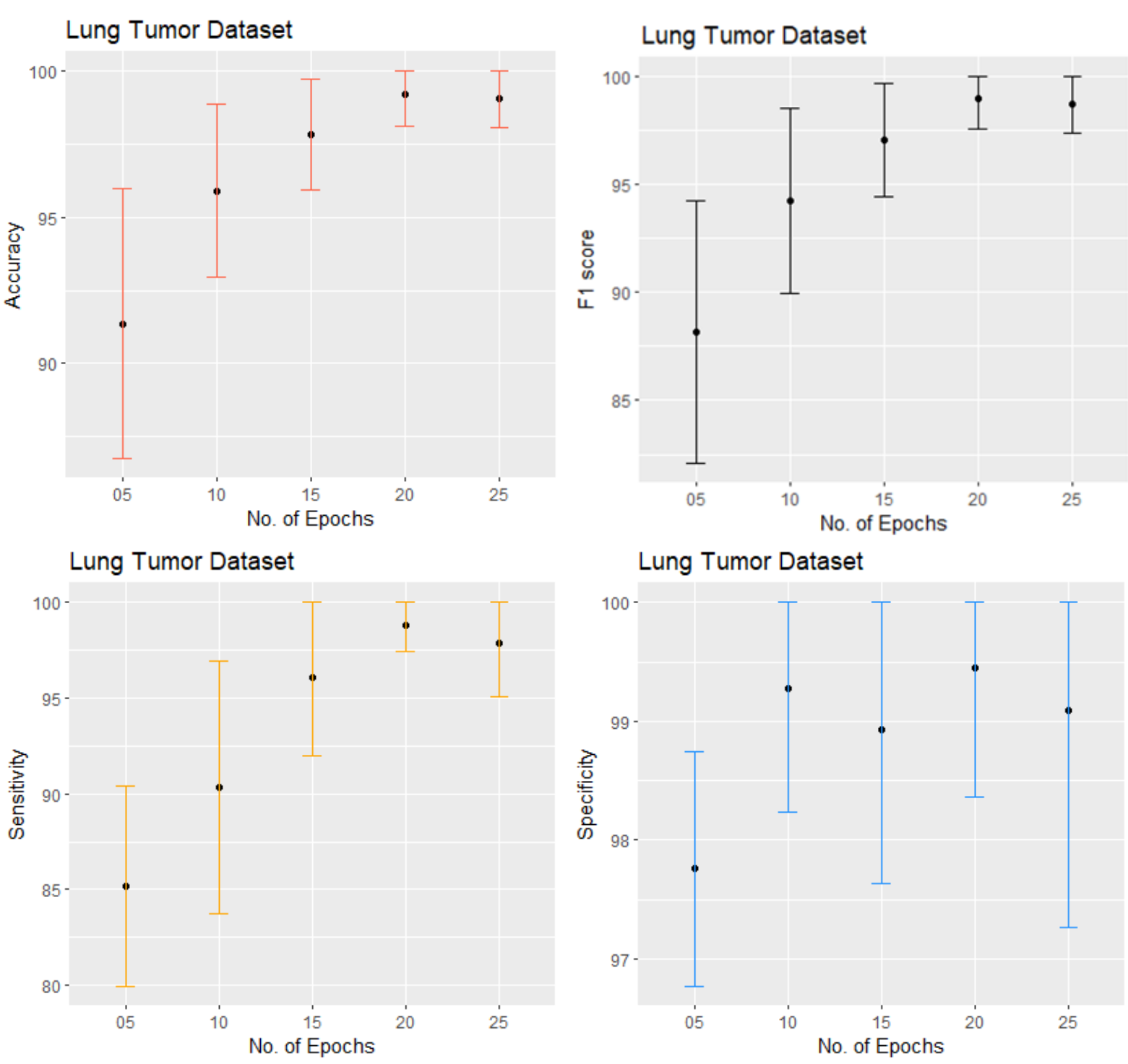

**Figure 8.** Lung tumor detection based on the kidney CT pretrained CNN model

### 3.5 Comparison of Performance with State-of-the-Art Pretrained DL Models

In this section, we assess the capacity of the proposed CNN architecture in transfer learning by comparing its classification performance against five widely used pretrained networks. Specifically, we evaluate the classification capabilities of Xception, VGG-16, VGG-19, MobileNetV2 and DenseNet121. Following the procedure outlined in Sections 3.2 to 3.4, we retrain these models for 20 epochs using a learning rate of $\eta = 10^{-4}$. The choice of 20 epochs is because this number generally yielded the best performance in most cases for the introduced CNN model.

Table 5 shows the best classification performance of the CNN models, which were retrained on the brain, lung and kidney cancer datasets. The top two performing pretrained models are highlighted in bold. The labels "Brain Pretrained", "Lung Pretrained" and "Kidney Pretrained" correspond to the introduced architecture, each pretrained for 75 epochs on the respective brain, lung and kidney datasets.

It is evident that the proposed computationally efficient CNN achieves the best tumor detection performance, regardless of the dataset used for pretraining. In nearly all experiments, the proposed CNN shows either the highest or second-highest performance. The only exception was the brain tumor dataset, where the VGG-16 model ranked second, achieving an accuracy of 91.17%, specificity of 84.33%, sensitivity of 98.00% and an F1-score of 91.73%. These results confirm that the proposed low-complexity model not only delivers satisfactory tumor detection performance but also outperforms several widely recognized pretrained CNN models [54].

**Table 5.** Comparison with widely used pretrained DL models

| Dataset | Model | Accuracy (%) | Specificity (%) | Sensitivity (%) | F1 score (%) |
|---|---|---|---|---|---|
| **Brain MRI scans** | Xception | 57.50 | 21.67 | 93.33 | 68.71 |
| | **VGG16** | **91.17** | **84.33** | **98.00** | **91.73** |
| | VGG19 | 89.67 | 84.67 | 94.67 | 90.16 |
| | MobileNetV2 | 78.67 | 80.33 | 77.00 | 78.31 |
| | DenseNet121 | 78.50 | 77.33 | 79.67 | 78.75 |
| | Lung Pretrained | 89.88 | 84.84 | 94.92 | 90.38 |
| | **Kidney Pretrained** | **92.54** | **91.02** | **94.87** | **92.71** |
| **Kidney CT scans** | Xception | 64.83 | 68.67 | 61.00 | 63.43 |
| | VGG16 | 99.67 | 99.33 | 100.00 | 99.67 |
| | VGG19 | 99.00 | 99.67 | 98.33 | 98.99 |
| | MobileNetV2 | 66.33 | 98.00 | 34.67 | 50.73 |
| | DenseNet121 | 87.00 | 81.00 | 93.00 | 87.74 |
| | **Brain Pretrained** | **99.96** | **99.92** | **100.00** | **99.91** |
| | **Lung Pretrained** | **99.87** | **99.75** | **100.00** | **99.87** |
| **Lung CT scans** | Xception | 66.36 | 99.27 | 12.05 | 21.28 |
| | VGG16 | 96.36 | 100.00 | 90.36 | 94.94 |
| | VGG19 | 98.18 | 99.27 | 96.39 | 97.56 |
| | MobileNetV2 | 64.55 | 97.08 | 10.84 | 18.75 |
| | DenseNet121 | 83.18 | 92.70 | 67.47 | 75.17 |
| | **Brain Pretrained** | **98.45** | **99.71** | **96.38** | **97.87** |
| | **Kidney Pretrained** | **99.18** | **99.56** | **98.55** | **99.10** |

## 4. Concluding Remarks

In this article, we introduce a computationally efficient CNN model for tumor detection based on both MRI and CT scans. The main objective of the present analysis is the investigation of the capacity of transfer learning based on the introduced low-complexity DL model. The reduced complexity of our network makes it well-suited for studying datasets with limited samples –common in medical studies, particularly medical imaging– compared to many approaches in the literature that rely on more complex and computationally expensive architectures.

Based on the above, the main conclusion of our study is that complex DL architectures are not always necessary to achieve high predictive performance. According to the task, comparable results can be obtained using more efficient, low-computational models which are robust against overfitting, easy to retrain and practical for users without access to extensive computational or storage resources.

Transfer learning offers significant advantages in medical imaging and tumor detection, particularly when dealing with limited labeled data, which is a common challenge in healthcare. Pretrained models often developed on large and diverse datasets, can be fine-tuned for specific tasks [52], such as detecting tumors through medical images. This approach allows models to leverage the rich feature representations learned from vast datasets, reducing the need for extensive annotated medical data, which can be costly and time-consuming to acquire. It also accelerates the training process, enhances model performance and improves generalization to unseen data [55], which is critical for efficient tumor detection. Overall, transfer learning empowers the development of robust and accurate diagnostic tools that can assist in early tumor detection and improve patient outcomes [31, 55].

According to the processing times reported in the Results section, the proposed architecture enables a fast training and testing process, which is also influenced by the input image size of 100×100 pixels, which is an adequate resolution in our case. However, it should be noted that the appropriate input size depends on the nature of the classification task, while the utilization of inputs of higher dimension may be requisite. Moreover, the observed decline in classification performance after a substantial increase in the number of epochs during fine-tuning is likely due to overfitting, which highlights the importance of systematically examining the number of epochs and considering early stopping strategies during fine-tuning.

We notice that the presented CNN scheme has shown significant efficiency regardless of the dataset utilized for the pretraining or fine-tuning phase, where only minor differentiations in the produced results are observed. On top of that, since the cardinality of the medical-related datasets is mostly limited, deep architectures would be prone to overfitting [52–53, 56–57]. The introduced low-complexity network overcomes this challenge and can be easily implemented under real-world circumstances, while its simplistic architecture encourages its retraining when new data is presented.

It should be noted that the objective of the present analysis is to evaluate the model's effectiveness across three distinct cancer types, each characterized by different morphological features, while also assessing the CNN's performance on both MRI and CT imaging modalities. The results indicate that the proposed architecture performs robustly across varying tasks and data types. Despite these promising findings, further experiments are necessary to evaluate the model's generalization to other biomedical imaging applications and rarer tumor types. Nevertheless, the retraining strategy employed in this study provides confidence in the model's adaptability with minimal computational cost. Finally, model interpretability remains an open challenge. Although the CNN demonstrates high accuracy, improving the transparency of its decision-making process, particularly in clinical contexts, would be essential to enhance trust and adoption among healthcare professionals.

For future work, several directions can be investigated to enhance the effectiveness of AI-driven cancer diagnosis systems. Incorporating more diverse and larger datasets, including multi-center data from varied demographic groups, could improve the model's generalizability, ensuring consistent performance across different populations. Additionally, integrating multimodal data, such as imaging combined with genomic, clinical, or electronic health record (EHR) information, could further enhance the model's predictive accuracy [58] and provide a more comprehensive understanding of cancer progression.

**Data Availability Statement**

The datasets used for the present analysis are included in public repositories, the URLs of which can be found in the Reference list.

**Conflict of Interest Statement**

The authors declare that they have no known competing financial interests or personal relationships that could have appeared to influence the work reported in this paper.

**Funding Statement**

This research has been supported by the EUropean Federation for CAncer Images (EUCAIM) project. Moreover, this study has been funded by the A.G. Leventis Foundation through a scholarship for doctoral studies.